\documentclass[10pt,twocolumn,letterpaper]{article}
\def\allfiles{}

\usepackage{cvpr}              

\usepackage{graphicx}
\usepackage{amsmath}
\usepackage{amssymb}
\usepackage{booktabs}
\makeatletter
\newcommand\footnoteref[1]{\protected@xdef\@thefnmark{\ref{#1}}\@footnotemark}
\makeatother
\usepackage[pagebackref,breaklinks,colorlinks]{hyperref}
\usepackage{float}

\usepackage[capitalize]{cleveref}
\crefname{section}{Sec.}{Secs.}
\Crefname{section}{Section}{Sections}
\Crefname{table}{Table}{Tables}
\crefname{table}{Tab.}{Tabs.}

\def\cvprPaperID{1315} 
\def\confName{CVPR}
\def\confYear{2022}

\begin{document}

\title{LGT-Net: Indoor Panoramic Room Layout Estimation with Geometry-Aware Transformer Network}

\author{Zhigang Jiang$^{1,2}$
\and
Zhongzheng Xiang$^{2}$
\and
Jinhua Xu$^{1}$\thanks{Corresponding author.}
\and
Ming Zhao$^{2}$
\and
$^1$East China Normal University
\qquad
$^2$Yiwo Technology\\
{\tt\small zigjiang@gmail.com\quad even\_and\_just@126.com \quad jhxu@cs.ecnu.edu.cn\quad zhaoming@123kanfang.com}\\
}

\maketitle

\begin{abstract}
\vspace{-2mm}
3D room layout estimation by a single panorama using deep neural networks has made great progress.
However, previous approaches can not obtain efficient geometry awareness of room layout with the only latitude of boundaries or horizon-depth.
We present that using horizon-depth along with room height can obtain omnidirectional-geometry awareness of room layout in both horizontal and vertical directions.
In addition, we propose a planar-geometry aware loss function with normals and gradients of normals to supervise the planeness of walls and turning of corners.
We propose an efficient network, LGT-Net, for room layout estimation, which contains a novel Transformer architecture called SWG-Transformer to model geometry relations.
SWG-Transformer consists of (Shifted) Window Blocks and Global Blocks to combine the local and global geometry relations. Moreover, we design a novel relative position embedding of Transformer to enhance the spatial identification ability for the panorama.
Experiments show that the proposed LGT-Net achieves better performance than current state-of-the-arts (SOTA) on benchmark datasets. The code is publicly available at~\url{https://github.com/zhigangjiang/LGT-Net}.
\end{abstract}
\vspace{-4mm}
\ifx\allfiles\undefined
\documentclass[10pt,twocolumn,letterpaper]{article}
\usepackage[review]{cvpr}      
\usepackage{graphicx}
\usepackage{amsmath}
\usepackage{amssymb}
\usepackage{booktabs}
\usepackage{makecell}
\usepackage{multirow}
\usepackage[pagebackref,breaklinks,colorlinks]{hyperref}
\usepackage[capitalize]{cleveref}
\crefname{section}{Sec.}{Secs.}
\Crefname{section}{Section}{Sections}
\Crefname{table}{Table}{Tables}
\crefname{table}{Tab.}{Tabs.}
\begin{document}
\fi
\def\inpufig{}
\section{Introduction}\label{sec:introduction}
\ifx\inpufig\undefined
\documentclass[10pt,twocolumn,letterpaper]{article}
\usepackage[review]{cvpr}      
\usepackage{graphicx}
\usepackage{amsmath}
\usepackage{amssymb}
\usepackage{booktabs}
\usepackage{makecell}
\usepackage{multirow}
\usepackage{CJKutf8}
\usepackage[pagebackref,breaklinks,colorlinks]{hyperref}
\usepackage[capitalize]{cleveref}
\crefname{section}{Sec.}{Secs.}
\Crefname{section}{Section}{Sections}
\Crefname{table}{Table}{Tables}
\crefname{table}{Tab.}{Tabs.}
\begin{document}
\fi


\begin{figure}
  \centering
  \includegraphics[width=1\linewidth]{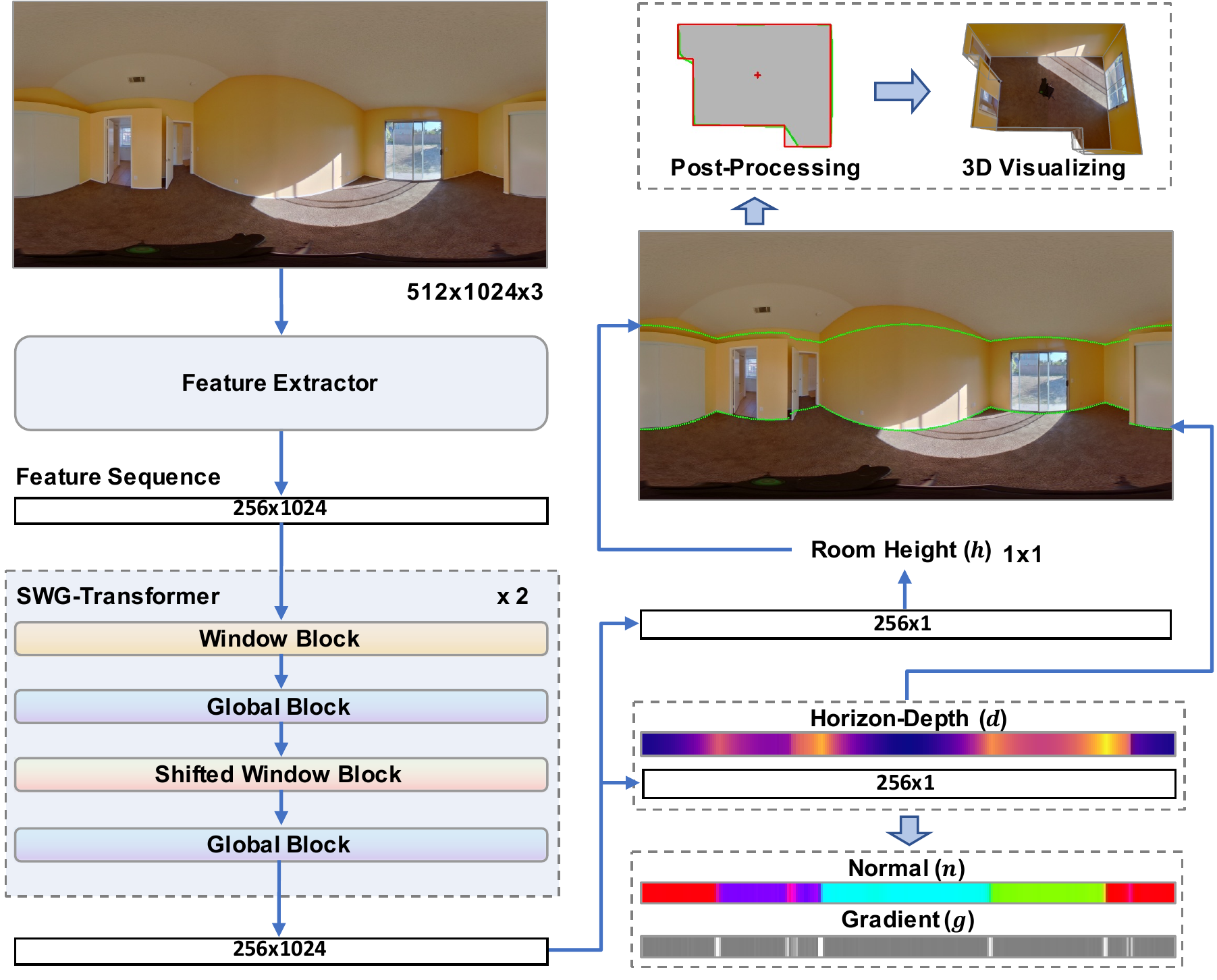}
  \caption{Overall architecture of the proposed LGT-Net. The network estimates the room layout from a single panorama using the \emph{omnidirectional}-geometry aware loss of horizon-depth and room height and the \emph{planar}-geometry aware loss of normals and gradients of normals. We visualize the predicted boundaries (green) by the horizon-depth and room height, and the floor plan (red) with post-processing by Manhattan constraint, finally output the 3D room layout.
  } 
  \label{fig:1_network}
  \vspace{-4mm}
\end{figure}
\ifx\inpufig\undefined
{\small
\bibliographystyle{ieee_fullname}
\bibliography{main}
}
\end{document}
\fi

The goal of estimating the 3D room layout by an indoor RGB image is to locate the corners or the floor-boundary and ceiling-boundary, as shown in \cref{fig:3_geometry_aware}, which plays a crucial role in 3D scene understanding\cite{sun2021hohonet}.
The panoramic images have wider (360$^\circ$) field of view (FoV) than perspective images and contain the whole-room contextual information\cite{zhang2014panocontext}.
With the development of deep neural networks and the popularity of panoramic cameras in recent years, 3D room layout estimation by a single panorama has made great achievements\cite{zou2018layoutnet, yang2019dula, horizon}.

Most room layouts conform to the Atlanta World assumption\cite{atlanta} with horizontal floor and ceiling, along with vertical walls\cite{atlantanet}. 
Thus the room layout can be represented by floor-boundary and room height, as shown in \cref{fig:3_geometry_aware}. 
However, previous approaches\cite{horizon, led, sun2021hohonet} estimate the room height by ceiling-boundary. 
And the networks predict the floor-boundary and ceiling-boundary with the same output branch, which affects each other since they need to predict both horizontal shape and vertical height of room layout.
Meanwhile, most previous approaches\cite{zou2018layoutnet, horizon, yang2019dula} use Manhattan constraint\cite{coughlan1999manhattan} or directly simplify boundaries\cite{atlantanet} in post-processing without considering the planar attribute of the walls to constrain the network output results.
In addition, for models \cite{horizon, led, sun2021hohonet} which formulate the room layout estimation task as 1D sequence prediction, a sequence processor is needed to model the  geometry relationship. Bidirectional Long Short-Term Memory (Bi-LSTM)\cite{schuster1997bidirectional, hochreiter1997long} is used in \cite{horizon, led}. Transformer\cite{vaswani2017attention} is an efficient framework for sequence processing and has made great success in natural language processing (NLP) tasks.
Vision Transformer (ViT)\cite{dosovitskiy2020image} has demonstrated strong abilities in the computer vision field recently. Nevertheless, there is no specially designed Transformer architecture for panoramas as we know.

Due to the above problems, we propose an efficient network called LGT-Net for panoramic room layout estimation. 
It contains a feature extractor to convert the panorama to feature sequence and a Transformer architecture as sequence processor.
Our proposed network directly predicts the room height and floor-boundary by two branches in the output layer, as shown in \cref{fig:1_network}.
Inspired by Wang \etal~\cite{led}, we represent the floor-boundary by horizon-depth. 
Thus, we propose an \emph{omnidirectional}-geometry aware loss function that computes the errors of horizon-depth and room height, which brings better geometry awareness of the room layout in both horizontal and vertical directions.
In addition, we observe the planar attribute of the walls and the turning attribute of the corners. Thus we propose to use the \emph{planar}-geometry aware loss function of normal consistencies and gradient of normal errors to supervise these attributes.

Moreover, we design a novel Transformer architecture called SWG-Transformer as the sequence processor for our network, which consists of (Shifted) Window Blocks\cite{liu2021Swin} and Global Blocks to combine the local and global geometry relations, as shown in \cref{fig:1_network}.
With the attention mechanism\cite{mnih2014recurrent}, our SWG-Transformer can better process the left and right borders of the panoramas than Bi-LSTM.
In addition, we design a novel relative position embedding\cite{raffel2020exploring, ke2020rethinking, shaw2018self} of Transformer architecture to enhance the spatial identification ability for the panoramas.

In order to demonstrate the effectiveness of our proposed approach, we conduct extensive experiments on benchmark datasets, including ZInD\cite{cruz2021zillow} dataset.
Meanwhile, we conduct ablation study on MatterportLayout\cite{zou2021manhattan} dataset in the following aspects: 
loss function, network architecture, and position embedding of Transformer to demonstrate the effectiveness of each component.
Experiments show that our proposed approach performs better than SOTA. The main contributions of our work are as follows:
\begin{itemize}
\vspace{-1mm}
\item We represent the room layout by horizon-depth and room height and output them with two branches of our network. Furthermore, we compute the horizon-depth and room height errors to form \emph{omnidirectional}-geometry aware loss function and compute normal and gradient errors to form \emph{planar}-geometry aware loss function.
\vspace{-1mm}
\item We show that exploiting Transfomer as a sequence processor is helpful for panoramic understanding. And our proposed  SWG-Transformer can better establish the local and global geometry relations of the room layout.
\vspace{-1mm}
\item We specially design a relative position embedding of Transformer to enhance the spatial identification ability for the panoramas.
\end{itemize}

\ifx\allfiles\undefined
{\small
\bibliographystyle{ieee_fullname}
\bibliography{main}
}
\end{document}
\fi
\ifx\allfiles\undefined
\documentclass[10pt,twocolumn,letterpaper]{article}
\usepackage[review]{cvpr}      
\usepackage{graphicx}
\usepackage{amsmath}
\usepackage{amssymb}
\usepackage{booktabs}
\usepackage{makecell}
\usepackage{multirow}
\usepackage{CJKutf8}
\usepackage[pagebackref,breaklinks,colorlinks]{hyperref}
\usepackage[capitalize]{cleveref}
\crefname{section}{Sec.}{Secs.}
\Crefname{section}{Section}{Sections}
\Crefname{table}{Table}{Tables}
\crefname{table}{Tab.}{Tabs.}
\makeatletter
\newcommand\footnoteref[1]{\protected@xdef\@thefnmark{\ref{#1}}\@footnotemark}
\makeatother

\begin{document}
\fi

\def\experiments{}
\section{Related Work}\label{sec:related_work}
\paragraph{Panoramic Room Layout Estimation}
Previous approaches mainly follow the Manhattan World assumption\cite{coughlan1999manhattan} or the less restrictive Atlanta World assumption\cite{atlanta} to estimate the room layout from a panorama and constrain post-processing.

Convolutional neural networks (CNNs) have been used to estimate the room layout with better performance.
Zou \etal~\cite{zou2018layoutnet} propose LayoutNet to predict probability maps of boundaries and corners and use layout parameter regressors to predict the final layout. 
Meanwhile, they extend the cuboid layout annotations of the Stanford\cite{armeni2017joint} dataset.
Yang \etal~\cite{yang2019dula} propose Dula-Net to predict floor and ceiling probability maps under both the equirectangular view and the perspective view of the ceiling.
Fernandez \etal~\cite{fernandez2020corners} propose to use equirectangular convolutions (EquiConvs) to estimate the room layout. 
Sun \etal~\cite{horizon} simplify the layout estimation task from 2D dense prediction to 1D sequence prediction.
They propose HorizonNet to extract the sequence by a feature extractor based on ResNet-50\cite{he2016deep}, 
then use Bi-LSTM as a sequence processor to establish the global relations.
We also use a framework composed of a feature extractor and a sequence processor.
Zou \etal~\cite{zou2021manhattan} propose improved version, LayoutNet v2 and Dula-Net v2, which have better performance on cuboid datasets than original approaches, and propose the general MatterportLayout dataset. 
However, their experiments show that HorizonNet\cite{horizon} is more efficient on general datasets. Pintore \etal~\cite{atlantanet} propose AtlantaNet to predict floor and ceiling boundary probability maps by same network instance and directly simplify \cite{douglas1973algorithms} output boundaries as post-processing.

Recently, Wang \etal~\cite{led} propose LE$\mathrm{D}^{2}$-Net\cite{led} to formulate the room layout estimation as predicting depth on the horizontal plane (horizon-depth), and they can pre-train on synthetic Structured3D\cite{zheng2020structured3d} dataset with deep information.
Sun \etal\cite{sun2021hohonet} propose HoHoNet to improve HorizonNet by re-designing the feature extractor with the Efficient Height Compression (EHC) module and employing multi-head self-attention (MSA)\cite{vaswani2017attention} as a sequence processor instead of Bi-LSTM.

\vspace{-4mm}
\paragraph{Geometry Awareness}
Wang \etal~\cite{led} propose a geometry-aware loss function of the room layout estimation by horizon-depth, which is only effective on horizontal direction.
Hu \etal~\cite{hu2019revisiting} propose to use losses of normal and gradient of depth to improve the performance for depth estimation on perspective images.
Eder \etal~\cite{eder2019pano} propose plane-aware loss that leverages curvature, normal, and point-to-plane distance to improve the performance for depth estimation on panoramic images. 
These works inspire us to propose a more effective geometry awareness loss function.

\vspace{-4mm}
\paragraph{Transformer}
Recently, ViT shows that Transformer architecture can compete with CNNs in visual classification tasks. 
Moreover, improved ViT networks (\eg, T2T-ViT\cite{Yuan_2021_ICCV}, PVT\cite{wang2021pyramid}, and Swin-Transformer\cite{liu2021Swin}) demonstrate that Transformer architecture is capable of surpassing CNNs.
Inspired by Swin-Transformer, we exploit window partition to reduce computation and enhance local modeling ability in SWG-Transformer. However, using window partition alone leads to lower global modeling ability.
Thus, our proposed SWG-Transformer consists of (Shifted) Window Blocks and Global Blocks to combine the local and global geometry relations.

\ifx\allfiles\undefined
{\small
\bibliographystyle{ieee_fullname}
\bibliography{main}
}
\end{document}
\fi
\ifx\allfiles\undefined
\documentclass[10pt,twocolumn,letterpaper]{article}
\usepackage[review]{cvpr}      
\usepackage{graphicx}
\usepackage{amsmath}
\usepackage{amssymb}
\usepackage{booktabs}
\usepackage[pagebackref,breaklinks,colorlinks]{hyperref}
\usepackage[capitalize]{cleveref}
\crefname{section}{Sec.}{Secs.}
\Crefname{section}{Section}{Sections}
\Crefname{table}{Table}{Tables}
\crefname{table}{Tab.}{Tabs.}
\DeclareMathOperator{\atantwo}{atan2}
\begin{document}
\fi

\def\inpufig{}
\section{Approach}
Our proposed approach aims to estimate the 3D room layout from a single panorama. 
We first describe the room layout representation with horizon-depth and room height and show that they can achieve \emph{omnidirectional}-geometry awareness (\cref{section:depth_ratio}).
Then, we introduce our proposed loss function, which consists of \emph{omnidirectional}-geometry aware loss and \emph{planar}-geometry aware loss (\cref{section:loss_function}).
Finally, we describe the network architecture of LGT-Net and use the SWG-Transformer to establish the local and global geometry relations of the room layout (\cref{section:network}).

\subsection{Panoramic Room Layout Representation}\label{section:depth_ratio}
\ifx\inpufig\undefined
\documentclass[10pt,twocolumn,letterpaper]{article}
\usepackage[review]{cvpr}      
\usepackage{graphicx}
\usepackage{amsmath}
\usepackage{amssymb}
\usepackage{booktabs}
\usepackage{makecell}
\usepackage{multirow}
\usepackage[pagebackref,breaklinks,colorlinks]{hyperref}
\usepackage[capitalize]{cleveref}
\crefname{section}{Sec.}{Secs.}
\Crefname{section}{Section}{Sections}
\Crefname{table}{Table}{Tables}
\crefname{table}{Tab.}{Tabs.}
\begin{document}
\fi


\begin{figure}
\centering
\includegraphics[width=1\linewidth]{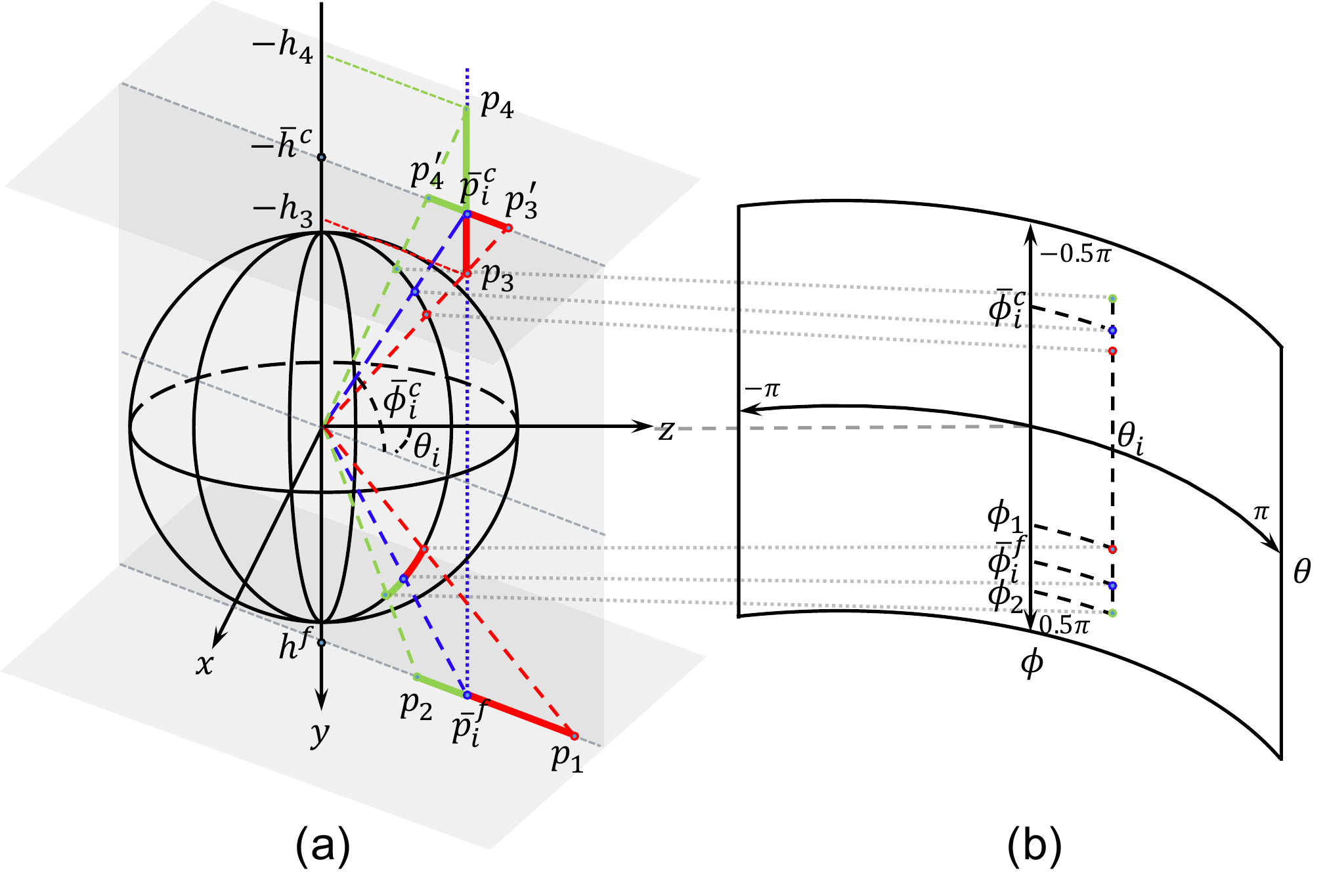}
\caption{
The mapping relationships between 3D points and a panorama.
\textbf{(a)} The coordinate relations in 3D space, where $h^{f}$ ($\bar{h}^{c}$) is the ground truth distance from camera center to the floor (ceiling).
\textbf{(b)} The longitude and latitude ($\theta, \phi$) relations on a panorama. 
}
\label{fig:3_mapping_relationship}
\vspace{-4mm}
\end{figure}

\ifx\inpufig\undefined
{\small
\bibliographystyle{ieee_fullname}
\bibliography{main}
}
\end{document}
\fi

We represent the room layout by the floor-boundary and room height, as shown in \cref{fig:3_geometry_aware}. 
We adopt a sampling approximation scheme to compute the floor-boundary. 
Specifically, sample $N$ points $\{ p_i \}_{i=1}^{N}$ with equal longitude intervals from the polygon of floor-boundary, where $N$ is $256$ by default in our implementation. The longitudes of the sampling points are denoted as $\{\theta_i = 2\pi(\frac{i}{N}- 0.5)\}_{i=1}^{N}$.
Then, we convert the points $\{ p_i \}_{i=1}^{N}$ to the horizon-depth sequence $\{d_i=D(p_i)\}_{i=1}^{N}$:
\vspace{-1mm}
\begin{equation}
\begin{aligned}   
p&=(x, y, z), \\
D(p)&= \sqrt{x^2 + z^2 }.
\end{aligned}
\label{eq:xz2depth}
\vspace{-1mm}
\end{equation}
Thus, we can estimate the room layout by predicting the horizon-depth sequence and room height. 

The floor-boundary on the ground plane is sensitive in the horizontal direction, as shown in \cref{fig:3_geometry_aware}.
HorizonNet\cite{horizon} predicts latitudes of ceiling and floor boundaries and calculates errors.
However, when the latitude errors of two sampling points are equal (\eg, $|\phi_1 - \bar\phi_i^f| = |\phi_2 - \bar\phi_i^f|$),
the corresponding horizon-depth errors may be different (\eg, $|D(p_1) - D(\bar{p}_i^f)| > |D(p_2) - D(\bar{p}_i^f)|$), as shown in \cref{fig:3_mapping_relationship}.
Thus, we predict horizon-depth and calculate errors to make better geometry awareness of the room layout in the horizontal direction.

Moreover, the room height is sensitive in the vertical direction, as shown in \cref{fig:3_geometry_aware}.
LE$\mathrm{D}^{2}$-Net\cite{led} also predicts latitudes but converts the latitudes of floor (ceiling) boundary to horizon-depth by projecting to ground truth floor (ceiling) plane to compute errors.
During inference, it calculates the room height by the consistency between the horizon-depth of ceiling and floor boundaries.
However, when the ceiling horizon-depth errors of the two sampling points are equal (\eg, $|D(p_{3}^{\prime}) - D(\bar{p}_i^c)| = |D(p_{4}^{\prime}) - D(\bar{p}_i^c)|$), 
the corresponding room height errors may be different (\eg, $p_{3}^{\prime}$ and $p_{4}^{\prime}$ are converted to $p_{3}$ and $p_{4}$ by the consistency of ground truth horizon-depth $D(\bar{p}_i^c)$, and  $|h_3 - \bar{h}^c| < |h_4 - \bar{h}^c|$), as shown in \cref{fig:3_mapping_relationship}.
Thus, we directly predict the room height and compute error to make better geometry awareness of the room layout in the vertical direction.

As a result, we propose an \emph{omnidirectional}-geometry aware loss function that computes the errors of horizon-depth and room height.
\cref{tab:ablation_study} shows the improvement in our approach. 

\subsection{Loss function}\label{section:loss_function}

\paragraph{Horizon-Depth and Room Height}
For the horizon-depth and room height, we apply the L1 loss:
\vspace{-1mm}
\begin{equation}
\begin{aligned}
&\mathcal{L}_d = \frac{1}{N} \sum_{i \in N} |d_i-\bar{d}_i|,\\
&\mathcal{L}_h = |h-\bar{h}|,
\label{eq:depth_ratio_loss}
\end{aligned}
\vspace{-1mm}
\end{equation}
where $\bar{d}_i$ ($\bar{h}$) is the ground truth  horizon-depth (room height), and $d_i$ ($h$) is the  predicted value.

\vspace{-4mm}
\paragraph{Normals}
As shown in \cref{fig:3_awareness}, each wall is a plane, but the positions on the same wall may have different horizon-depth (\eg, $D(p_{i-1})\neq D(p_{i})$). 
However, the normals at different positions on the same wall plane are consistent.
Thus we use normal consistency to supervise the planar attribute of the walls.
Specifically, when the network predicts the horizon-depth sequence $\{d_i
\}_{i=1}^{N}$, we convert each horizon-depth $d_i$ to the corresponding 3D point $p_i$ and obtain the normal vector $n_i$ that is always perpendicular to the $y$-axis.
Then we compute the cosine similarity to get the normal loss:
\vspace{-1mm}
\begin{equation}
\begin{aligned}
p_i &= (d_i \sin(\theta_i), ~h^f, ~d_i\cos(\theta_i)),\\
n_i &= M_r(\frac{p_{i+1} - p_i}{\left\| {p_{i+1} - p_i}\right\|})^T,\\
\mathcal{L}_n &= \frac{1}{N} \sum_{i \in N}(- n_i \cdot \bar{n}_i) ,
\label{eq:normal_loss}
\end{aligned}
\vspace{-1mm}
\end{equation}
where $ M_r$ is the rotation matrix of $\frac{\pi}{2}$, $\bar n_i$ is the ground truth of normal vector, and $n_i$ is the predicted normal vector.

\vspace{-4mm}
\paragraph{Gradients of normals}
The normals change near the corners, as shown in \cref{fig:3_plane_awareness}.
In order to supervise the turning of corners, we compute the angle between $n_{i-1}$ and $n_{i+1}$ to represent gradient $g_i$ of normal angle, then apply the L1 loss:
\vspace{-1mm}
\begin{equation}
\begin{aligned}
g_i &= \arccos(n_{i-1} \cdot n_{i+1}),\\
\mathcal{L}_g &= \frac{1}{N} \sum_{i \in N} |g_i-\bar{g}_i|,
\label{eq:gradient_loss}
\end{aligned}
\vspace{-1mm}
\end{equation}
where $\bar{g}_i$ and $g_i$ are the ground truth and predicted gradients, respectively.

\vspace{-4mm}
\paragraph{Total Loss}

\ifx\inpufig\undefined
\documentclass[10pt,twocolumn,letterpaper]{article}
\usepackage[review]{cvpr}      
\usepackage{graphicx}
\usepackage{amsmath}
\usepackage{amssymb}
\usepackage{booktabs}
\usepackage{makecell}
\usepackage{multirow}
\usepackage[pagebackref,breaklinks,colorlinks]{hyperref}
\usepackage[capitalize]{cleveref}
\crefname{section}{Sec.}{Secs.}
\Crefname{section}{Section}{Sections}
\Crefname{table}{Table}{Tables}
\crefname{table}{Tab.}{Tabs.}
\begin{document}
\fi


\begin{figure}
  \centering
  \begin{subfigure}{0.5\linewidth}
  \raisebox{-1.165\height}{\includegraphics[width=1\columnwidth]{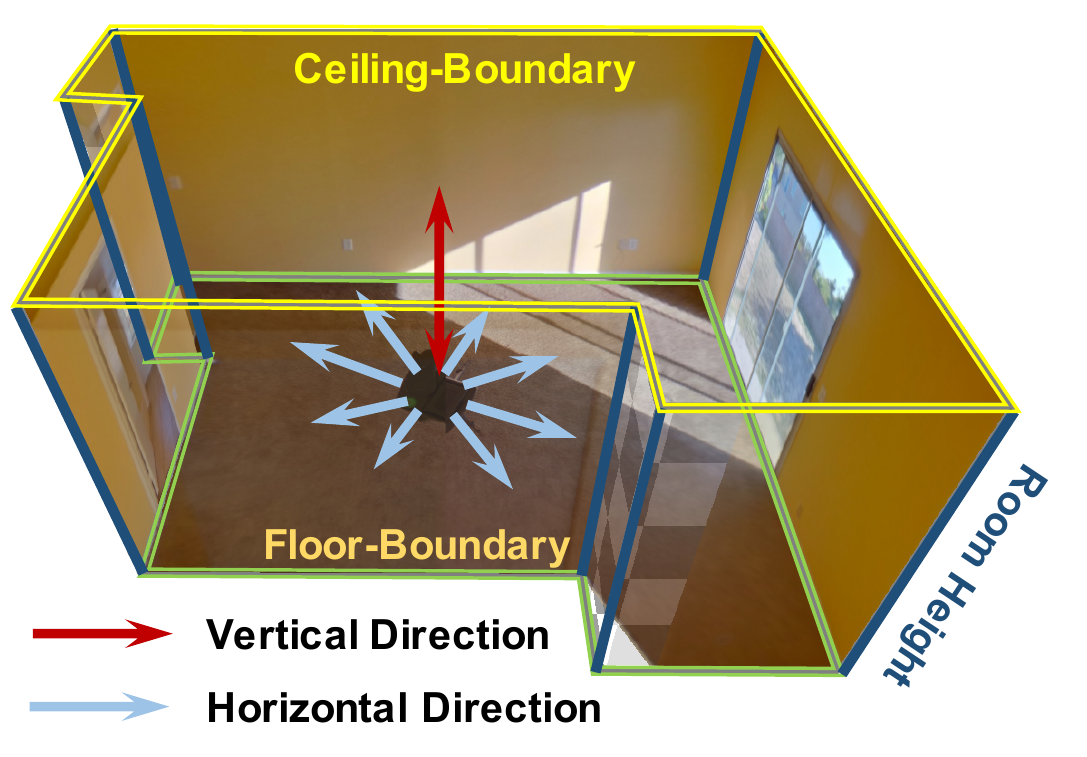}}
    \caption{Omnidirectionl Geometry Awareness.}
    \label{fig:3_geometry_aware}
  \end{subfigure}
  \begin{subfigure}{0.49\linewidth}
    \includegraphics[width=1\linewidth]{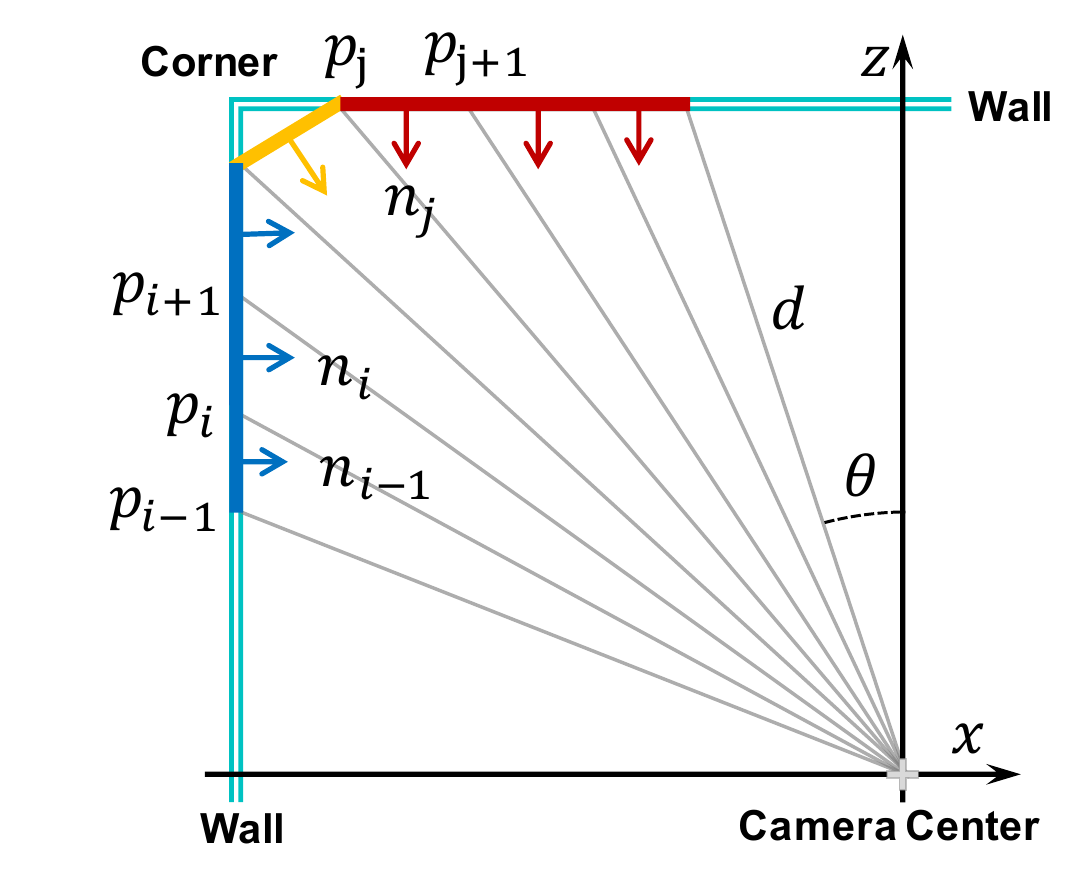}
    \caption{Planar Geometry Awareness.}
    \label{fig:3_plane_awareness}
  \end{subfigure}
  \caption{Illustration of geometry awareness for the room layout. \textbf{(a)} The horizontal and vertical directions influence the room layout. We propose \emph{omnidirectional}-geometry aware loss function by horizon-depth and room height. \textbf{(b)} The \emph{planar}-geometry awareness by normals.}
  \label{fig:3_awareness}
  \vspace{-4mm}
\end{figure}
\ifx\inpufig\undefined
{\small
\bibliographystyle{ieee_fullname}
\bibliography{main}
}
\end{document}
\fi

The loss terms related to horizon-depth and room height enhance the \emph{omnidirectional}-geometry awareness. And the loss terms corresponding to the normals and gradients enhance the \emph{planar}-geometry awareness.
Therefore, to enhance both aspects, we use a total loss function as follows:
\vspace{-1mm}
\begin{equation}
\mathcal{L}=\lambda\mathcal{L}_d+\mu\mathcal{L}_h+\nu(\mathcal{L}_n+\mathcal{L}_g),
\label{eq:total_loss}
\vspace{-1mm}
\end{equation}
where $\lambda, \mu, \nu \in \mathbb{R}$ are hyper-parameters to balance the contribution of each component loss.

\subsection{Network}\label{section:network}
\ifx\inpufig\undefined
\documentclass[10pt,twocolumn,letterpaper]{article}
\usepackage[review]{cvpr}      
\usepackage{graphicx}
\usepackage{amsmath}
\usepackage{amssymb}
\usepackage{booktabs}
\usepackage{makecell}
\usepackage{multirow}
\usepackage[pagebackref,breaklinks,colorlinks]{hyperref}
\usepackage[capitalize]{cleveref}
\crefname{section}{Sec.}{Secs.}
\Crefname{section}{Section}{Sections}
\Crefname{table}{Table}{Tables}
\crefname{table}{Tab.}{Tabs.}
\begin{document}
\fi


\begin{figure*}
  \centering
  \includegraphics[width=1\linewidth]{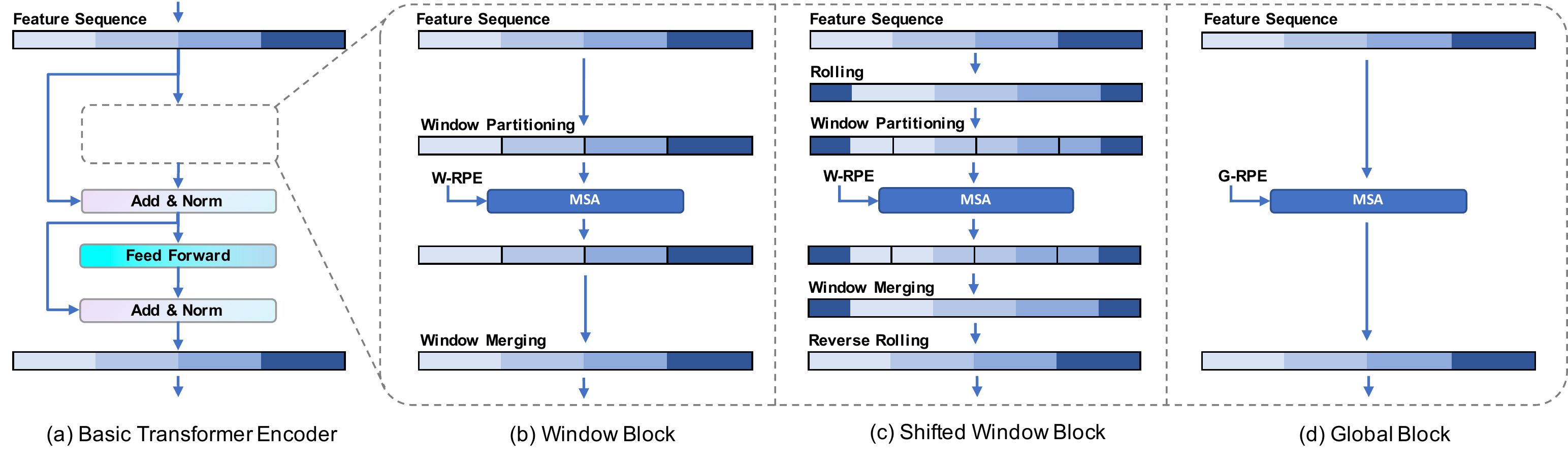}
  \caption{Illustration of SWG-Transformer Blocks. W-RPE and G-RPE are integrated into MSA for each block.
  \textbf{(a)} All blocks are based on the original Transformer\cite{vaswani2017attention} encoder.
  \textbf{(b)} Window Block needs to partition and merge windows before and after MSA.
  \textbf{(c)} Shifted Window Block needs to roll and reverse roll sequence feature before and after Window Block operation.
  \textbf{(d)} Global Block does not add additional operations.
  }
  \label{fig:3_swgblocks}
  \vspace{-2mm}
\end{figure*}

\ifx\inpufig\undefined
{\small
\bibliographystyle{ieee_fullname}
\bibliography{main}
}
\end{document}
\fi

Our proposed LGT-Net consists of a feature extractor and a sequence processor, as shown in \cref{fig:1_network}.
The feature extractor extracts a feature sequence from a panorama. 
Then, our proposed SWG-Transformer processes the  feature sequence.
In the end, our network respectively predicts the horizon-depth sequence and a room height value by two branches in the output layer.

\vspace{-4mm}
\paragraph{Feature Extractor}
In our implementation, the feature extractor uses the architecture proposed in HorizonNet \cite{horizon} based on ResNet-50\cite{he2016deep}.
The architecture takes a panorama with dimension of $512 \times 1024 \times 3$ (height, width, channel) as input and gets 2D feature maps of $4$ different scales by ResNet-50.
Then, it compresses the height and up samples width $N$ of each feature map to get 1D feature sequences with same dimension $\mathbb{R}^{N \times \frac{D}{4}}$ and connect them, finally outputs a feature sequence $\mathbb{R}^{N \times D}$, where $D$ is $1024$ in our implementation.
Moreover, we can also use the EHC module proposed by Sun \etal\cite{sun2021hohonet} or Patch Embedding\cite{dosovitskiy2020image} of ViT\cite{dosovitskiy2020image} (described in \cref{section:general_results}) as the feature extractor to extract the feature sequence.

\vspace{-4mm}
\paragraph{SWG-Transformer}
In our proposed SWG-Transformer, each loop contains four successive blocks, in the following order: Window Block, Global Block, Shifted Window Block, Global Block. 
The default loop is repeated twice ($\times 2$) for a total of 8 blocks, as shown in \cref{fig:1_network}.
Each block follows the basic Transformer\cite{vaswani2017attention} encoder architecture, as shown in \cref{fig:3_swgblocks}a, and the difference lies in the operations before and after MSA.
Moreover, the dimension of the sequence and corresponding positions of tokens are the same in the input sequence and output sequence of each block.

In Window Block, we use window partition for the input feature sequence and get $\frac{N}{N_w}$ window feature sequences $\mathbb{R}^{ N_w \times D}$ before MSA, where $N_w$ denotes the window length and is set to $16$ by default in our implementation.
The window partition enhances local geometry relations and reduces the computation when calculating self-attention. 
Moreover, the window feature sequences are merged after the MSA, as shown in \cref{fig:3_swgblocks}b.

Shifted Window Block aims to connect adjacent windows to enhance information interaction, and it is based on the Window Block.
We roll the input feature sequence with $\frac{N_w}{2}$ as its offset before the window partition.
To restore the original positions of feature sequence after merging the window feature sequences, we perform a reverse roll operation, as shown in \cref{fig:3_swgblocks}c.

In Global Window Block, operations like window partitioning and rolling are unnecessary. It follows the original Transformer\cite{vaswani2017attention} encoder architecture and aims to enhance the global geometry relations, as shown in \cref{fig:3_swgblocks}d.

\vspace{-4mm}
\paragraph{Position Embedding}

\ifx\inpufig\undefined
\documentclass[10pt,twocolumn,letterpaper]{article}
\usepackage[review]{cvpr}      
\usepackage{graphicx}
\usepackage{amsmath}
\usepackage{amssymb}
\usepackage{booktabs}
\usepackage{makecell}
\usepackage{multirow}
\usepackage[pagebackref,breaklinks,colorlinks]{hyperref}
\usepackage[capitalize]{cleveref}
\crefname{section}{Sec.}{Secs.}
\Crefname{section}{Section}{Sections}
\Crefname{table}{Table}{Tables}
\crefname{table}{Tab.}{Tabs.}
\begin{document}
\fi


\begin{figure}
  \centering
  \begin{subfigure}{0.5\linewidth}
    \includegraphics[width=1\linewidth]{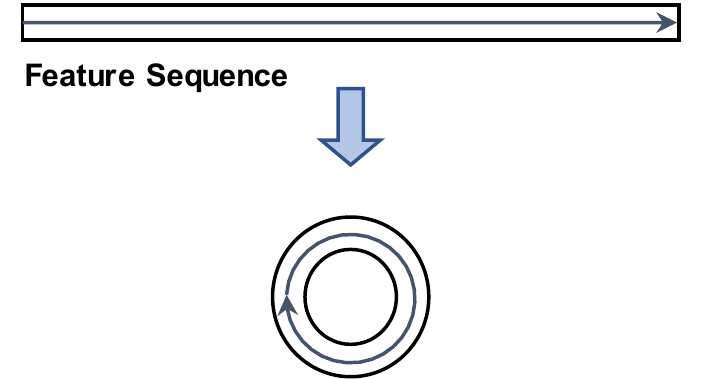}
    \caption{}
    \label{fig:3_rpe_ring}
  \end{subfigure}
  \hspace{4mm}
  \begin{subfigure}{0.27\linewidth}
    \includegraphics[width=1\linewidth]{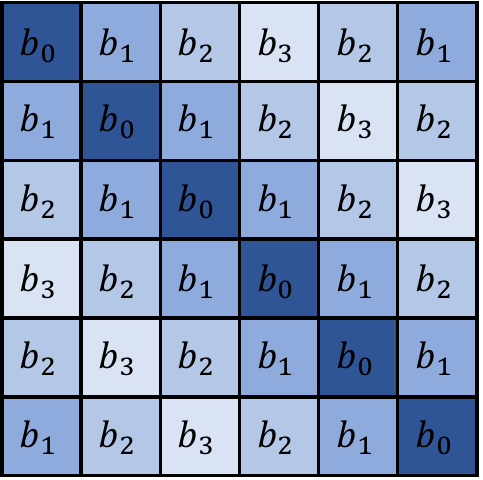}
    \caption{}
    \label{fig:3_rpe_bias}
  \end{subfigure}
  \caption{
  \textbf{(a)}The feature sequence from panorama is a circular structure.
  \textbf{(b)} Illustration of relative position bias matrix for Global Block.}
  \label{fig:3_rpe}
  \vspace{-4mm}
\end{figure}

\ifx\inpufig\undefined
{\small
\bibliographystyle{ieee_fullname}
\bibliography{main}
}
\end{document}
\fi
Since the pure attention module is insensitive to positions of distinguishing tokens, the spatial recognition ability is weakened.
Thus, in computing self-attention, we use the relative position embedding of T5\cite{raffel2020exploring} to enhance the spatial identification ability.
Specifically, we denote the input sequence of MSA as $X=\{ x_i \}_{i=1}^{M}$, where $M$ is the sequence length and $x_i \in \mathbb{R}^D$. A bias matrix $B \in \mathbb{R}^{M \times M}$ is added to Scaled Query-Key product\cite{vaswani2017attention}:
\begin{equation}
\begin{aligned}
&\alpha_{ij}=\frac{1}{\sqrt{D}}(x_{i} W^{Q})(x_{j} W^{K})^{T}+B_{ij},\\
&\text{Attention}(X)=\text{Softmax}(\alpha)(XW^{V}),
\end{aligned}
\end{equation}
where $W^Q, W^K, W^V \in \mathbb{R}^{D \times D}$ are learnable project matrices, each bias $B_{ij}$ comes from a learnable scalar table.

In (Shifted) Window Block, $M=N_w$.
We denote the learnable scalar table as $\{ b_k \}_{k=-N_w+1}^{N_w-1}$, and $B_{ij}$ corresponds to $b_{j-i}$. 
This scheme is denoted as W-RPE and integrated into MSA, as shown in \cref{fig:3_swgblocks}b and \cref{fig:3_swgblocks}c.

In Global Block, $M=N$.
As shown in \cref{fig:3_rpe_ring}, the feature sequence is a circular structure. 
If we use a scheme similar to Window Block and denote the learnable scalar table as $\{ b_k \}_{k=-N+1}^{N-1}$, it will result in the same distance represented twice from different directions. 
Specifically, $B_{ij}$ corresponds to $b_{j-i}$ and also corresponds to $b_{j-N-i}$.
Thus, we propose a \emph{symmetric} representation of only distance and denote the learnable scalar table as $\{ b_k \}_{k=0}^{n}$, where $n=\frac{N}{2}$.
When $ |j-i|\leq \frac{N}{2}$, $B_{ij}$ corresponds to $b_{|j-i|}$, otherwise $B_{ij}$ corresponds to $b_{N-|j-i|}$. 
A visualization of bias matrix is shown in \cref{fig:3_rpe_bias}.
We denote this scheme as G-RPE and integrate it into MSA, as shown in \cref{fig:3_swgblocks}d.
\ifx\allfiles\undefined
{\small
\bibliographystyle{ieee_fullname}
\bibliography{main}
}
\end{document}
\fi
\ifx\allfiles\undefined
\documentclass[10pt,twocolumn,letterpaper]{article}
\usepackage[review]{cvpr}      
\usepackage{graphicx}
\usepackage{amsmath}
\usepackage{amssymb}
\usepackage{booktabs}
\usepackage{makecell}
\usepackage{multirow}
\usepackage{CJKutf8}
\usepackage[pagebackref,breaklinks,colorlinks]{hyperref}
\usepackage[capitalize]{cleveref}
\crefname{section}{Sec.}{Secs.}
\Crefname{section}{Section}{Sections}
\Crefname{table}{Table}{Tables}
\crefname{table}{Tab.}{Tabs.}
\makeatletter
\newcommand\footnoteref[1]{\protected@xdef\@thefnmark{\ref{#1}}\@footnotemark}
\makeatother

\begin{document}
\fi

\def\experiments{}
\section{Experiments}

We implement LGT-Net using PyTorch\cite{paszke2019pytorch} and use the Adam optimizer\cite{DBLP:journals/corr/KingmaB14} with $\beta_1=0.9$, $\beta_2=0.999$, and the learning rate is set to $0.0001$. 
We train the network on a single NVIDIA GTX 1080 Ti GPU for 200 epochs on ZInD\cite{cruz2021zillow} dataset and 1000 epochs on other datasets, with batch size of 6.
We adopt the same data augmentation approaches mentioned in Horizon-Net\cite{horizon}, including standard left-right flipping, panoramic horizontal rotation, luminance change, and pano stretch during training. 
In addition, we set hyper-parameters as $\lambda=0.9, \mu=0.1, \nu=0.1$ in \cref{eq:total_loss}.

\subsection{Datasets}\label{section:datasets}
\paragraph{PanoContext and Stanford 2D-3D}
PanoContext\cite{zhang2014panocontext} dataset contains 514 annotated cuboid room layouts. 
Stanford 2D-3D\cite{armeni2017joint} dataset  contains 552 cuboid room layouts labeled by Zou \etal~\cite{zou2018layoutnet} and has a smaller vertical FoV than other datasets.
We follow the same training/validation/test splits of LayoutNet\cite{zou2018layoutnet} to evaluate these two datasets.

\vspace{-4mm}
\paragraph{MatterportLayout}
MatterportLayout\cite{zou2021manhattan} dataset  is a subset of Matterport3D\cite{chang2017matterport3d} dataset. 
It contains 2,295 general room layouts labeled by Zou \etal~\cite{zou2021manhattan}.
We follow the same training/validation/test splits for evaluation.

\vspace{-4mm}
\paragraph{ZInD}
To the best of our knowledge, ZInD\cite{cruz2021zillow} dataset is currently the largest dataset with room layout annotations. 
It better mimics the real-world data distribution since it includes cuboid, more general Manhattan, non-Manhattan, and non-flat ceilings layouts.
ZInD\cite{cruz2021zillow} dataset contains $67448$ panoramas from $1575$ real unfurnished residential homes\footnote{\url{https://github.com/zillow/zind}} and separates a ``simple'' subset that every room layout does not have any contiguous occluded corners.
We experiment on the ``simple" subset 
and use the ``raw'' layout annotations, 
and follow the official training/validation/test splits at the per-home level. 
In addition, we filter 0.8\% of layout annotations that do not contain the camera center. 
In total, we have the training, validation, and test splits consisting of $24882$, $3080$, and $3170$ panoramas, respectively.

\subsection{Evaluation Metrics}
We use the standard evaluation metrics proposed by Zou \etal~\cite{zou2018layoutnet}: intersection over union of floor shapes (2DIoU) and 3D room layouts (3DIoU), corner error (CE), and pixel error (PE). Meanwhile, we evaluate the depth accuracy with root mean squared error (RMSE) by using the camera height of $1.6$ meters and the percentage of pixels ($\delta_{1}$) where the ratio between prediction depth and ground truth depth is within a threshold of $1.25$ mentioned in Zou \etal~\cite{zou2021manhattan}.

\subsection{Cuboid Room Results}\label{section:cuboid_results}
\ifx\experiments\undefined
\documentclass[10pt,twocolumn,letterpaper]{article}
\usepackage[review]{cvpr}      
\usepackage{graphicx}
\usepackage{amsmath}
\usepackage{amssymb}
\usepackage{booktabs}
\usepackage{makecell}
\usepackage{multirow}
\usepackage{CJKutf8}
\usepackage[pagebackref,breaklinks,colorlinks]{hyperref}
\usepackage[capitalize]{cleveref}
\crefname{section}{Sec.}{Secs.}
\Crefname{section}{Section}{Sections}
\Crefname{table}{Table}{Tables}
\crefname{table}{Tab.}{Tabs.}
\begin{document}
\fi


\begin{table}
\small
\begin{tabular*}{\hsize}{@{}@{\extracolsep{\fill}}lccc@{}}
\toprule
Method  &3DIoU(\%)  &CE(\%)  &PE(\%)  \\
\midrule
\multicolumn{4}{c}{Train on PanoContext + Whole Stnfd.2D3D datasets } \\
\midrule
LayoutNet v2~\cite{zou2021manhattan} & 85.02 & \textbf{0.63} & \textbf{1.79}\\
DuLa-Net v2~\cite{zou2021manhattan} & 83.77 & 0.81 & 2.43\\
HorizonNet ~\cite{horizon} & 82.63 & 0.74 & 2.17\\
Ours & \textbf{85.16} & - & - \\
Ours [w/ Post-proc] & 84.94 & 0.69 & 2.07 \\
\midrule
\multicolumn{4}{c}{Train on Stnfd.2D3D + Whole PanoContext datasets } \\
\midrule
LayoutNet v2~\cite{zou2021manhattan} & 82.66 & 0.83 & 2.59\\
DuLa-Net v2~\cite{zou2021manhattan} & \textbf{86.60} & 0.67 & 2.48\\
HorizonNet ~\cite{horizon} & 82.72 & 0.69 & 2.27\\
AtlantaNet ~\cite{atlantanet} & 83.94 & 0.71 & 2.18\\
Ours & 85.76 & - & - \\
Ours [w/ Post-proc] & 86.03 & \textbf{0.63} & \textbf{2.11} \\
\bottomrule
\end{tabular*}
\caption{Quantitative results of cuboid layout estimation evaluated on PaonContext\cite{zhang2014panocontext} (top) and Stanford 2D–3D\cite{armeni2017joint} (bottom) datasets.}
\label{tab:cuboid}
\vspace{-4mm}
\end{table}

\ifx\experiments\undefined
{\small
\bibliographystyle{ieee_fullname}
\bibliography{main}
}
\end{document}
\fi

Since data in a single dataset is limited, it may lead to bias.
We use a combined dataset scheme mentioned in Zou \etal~ \cite{zou2021manhattan} for training.
The combined dataset contains a training split of the current evaluation dataset and another whole dataset. 
We provide the quantitative results of the cuboid layout in \cref{tab:cuboid}.
In addition, some baseline results include post-processing.
We also report results with a post-processing of DuLa-Net\cite{yang2019dula} (denoted as ``Ours [w/ Post-proc]"). 
Meanwhile, CE and PE values are reported.

\vspace{-4mm}
\paragraph{PanoContext}
LayoutNet v2\cite{zou2021manhattan} gives slightly better CE and PE performance than ours. 
And we argue that its 2D convolution for corner location and the post-processing method of gradient ascent is more effective for cuboid layouts.
However, our approach offers better performance than all the other SOTA approaches with respect to 3DIoU.

\vspace{-4mm}
\paragraph{Stanford 2D-3D} 
Dula-Net v2\cite{zou2021manhattan} gives slightly better 3DIoU than ours, and we argue that it uses perspective view, which is more effective for panoramas with small vertical FoV.
However, our approach offers better performance than similar approaches\cite{horizon, led} predicting on equirectangular view.

\subsection{General Room Results}\label{section:general_results}
\ifx\experiments\undefined
\documentclass[10pt,twocolumn,letterpaper]{article}
\usepackage[review]{cvpr}      
\usepackage{graphicx}
\usepackage{amsmath}
\usepackage{amssymb}
\usepackage{booktabs}
\usepackage{makecell}
\usepackage{multirow}
\usepackage{CJKutf8}
\usepackage[pagebackref,breaklinks,colorlinks]{hyperref}
\usepackage[capitalize]{cleveref}
\crefname{section}{Sec.}{Secs.}
\Crefname{section}{Section}{Sections}
\Crefname{table}{Table}{Tables}
\crefname{table}{Tab.}{Tabs.}
\begin{document}
\fi


\begin{figure*}
  \centering
  \begin{subfigure}{1\linewidth}
    \includegraphics[width=1\linewidth]{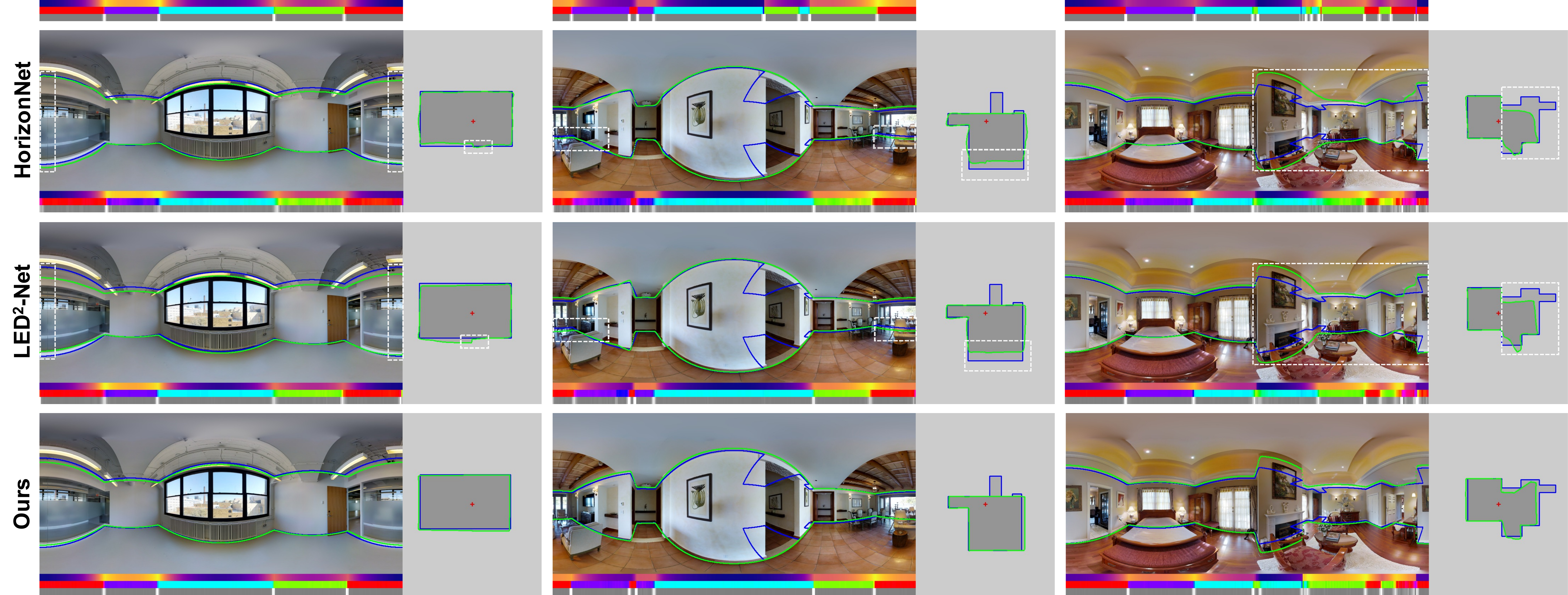}
    \caption{Qualitative comparison on MatterportLayout\cite{zou2021manhattan} dataset.}
    \label{fig:4_qualitative_comparison_mp3d}
  \end{subfigure}
  \\
  \vspace{2mm}
  \begin{subfigure}{1\linewidth}
    \includegraphics[width=1\linewidth]{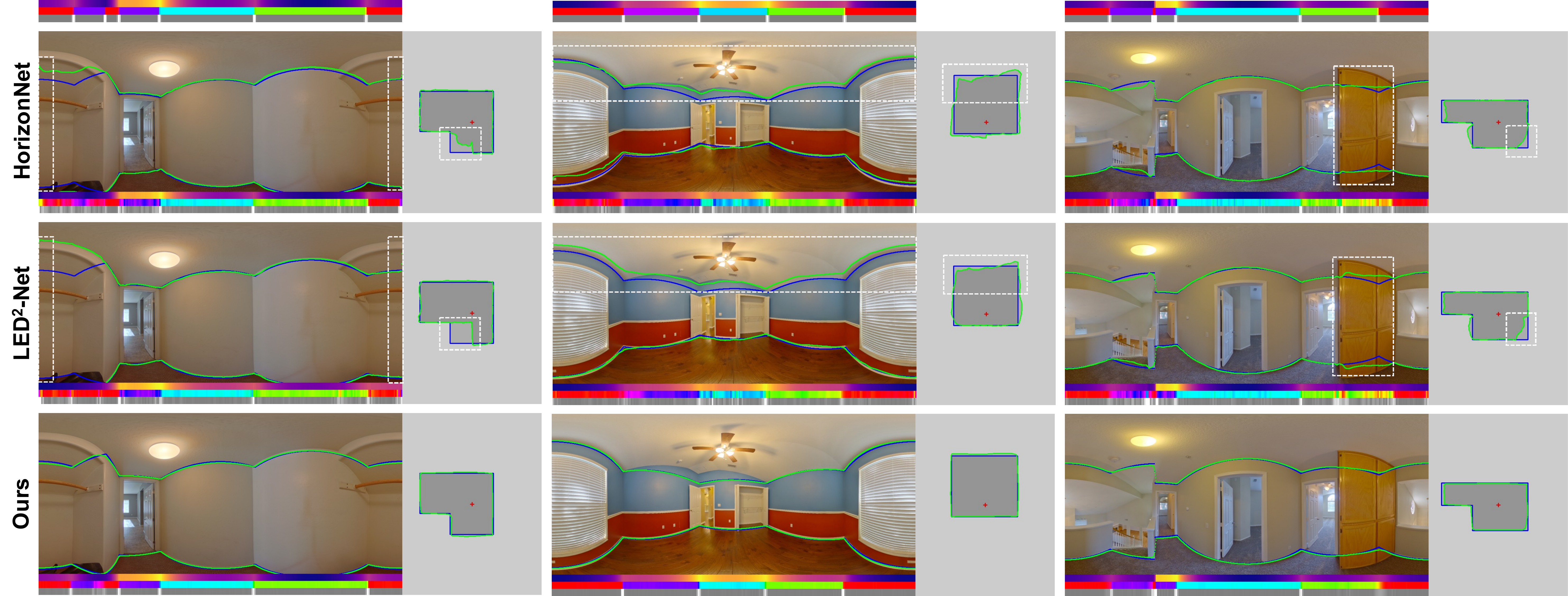}
    \setlength{\belowcaptionskip}{-2mm}\centering\caption{Qualitative comparison on ZInd\cite{cruz2021zillow} dataset.}
    \label{fig:4_qualitative_comparison_zind}
  \end{subfigure}
  \caption{
  Qualitative comparison of general layout estimation. We show the room layouts without post-processing by HorizonNet\cite{horizon}, LE$\mathrm{D}^{2}$-Net\cite{led}, and ours on MatterportLayout\cite{zou2021manhattan} dataset (top) and  ZInd\cite{cruz2021zillow} dataset (bottom). We show the boundaries of room layout on panorama (left) and the floor plan (right). The blue lines are ground truth, and the green lines are prediction. Moreover, we visualize the predicted horizon-depth, normal, and gradient below each panorama and the ground truth in the first row. The dashed white lines highlight the errors generated by the baselines.}
  \label{fig:4_qualitative_comparison}
   \vspace{-4mm}
\end{figure*}
\ifx\experiments\undefined
{\small
\bibliographystyle{ieee_fullname}
\bibliography{main}
}
\end{document}
\fi

\ifx\experiments\undefined
\documentclass[10pt,twocolumn,letterpaper]{article}
\usepackage[review]{cvpr}      
\usepackage{graphicx}
\usepackage{amsmath}
\usepackage{amssymb}
\usepackage{booktabs}
\usepackage{makecell}
\usepackage{multirow}
\usepackage{CJKutf8}
\usepackage[pagebackref,breaklinks,colorlinks]{hyperref}
\usepackage[capitalize]{cleveref}
\crefname{section}{Sec.}{Secs.}
\Crefname{section}{Section}{Sections}
\Crefname{table}{Table}{Tables}
\crefname{table}{Tab.}{Tabs.}
\begin{document}
\fi


\begin{table}
\small
\begin{tabular*}{\hsize}{@{}@{\extracolsep{\fill}}p{2.60cm}p{1cm}p{1cm}p{0.50cm}p{0.7cm}@{}}
\toprule
Method&2DIoU(\%) &3DIoU(\%) & RMSE & $\delta_{1}$\\
\midrule
LayoutNet v2~\cite{zou2021manhattan} & 78.73 & 75.82 & 0.258 & 0.871    \\
DuLa-Net v2~\cite{zou2021manhattan} & 78.82 & 75.05 & 0.291 & 0.818 \\
HorizonNet~\cite{horizon} & 81.71 & 79.11 & \textbf{0.197} & 0.929	 \\
AtlantaNet~\cite{atlantanet} & 82.09 & 80.02 & - & - \\
HoHoNet~\cite{sun2021hohonet} & 82.32 & 79.88 & - & - \\
LE$\mathrm{D}^{2}$-Net~\cite{led} & 82.61 & 80.14 & 0.207 & 0.947 \\
Ours & \textbf{83.52} & \textbf{81.11} & 0.204 & \textbf{0.951}\\
Ours [w/ Post-proc] & \textbf{83.48} & \textbf{81.08} & 0.214 & 0.940 \\
\bottomrule
\end{tabular*}
\caption{Quantitative results of general layout estimation evaluated on MatterportLayout\cite{zou2021manhattan} dataset.
}
\label{tab:matterport}
\vspace{-2mm}
\end{table}

\ifx\experiments\undefined
{\small
\bibliographystyle{ieee_fullname}
\bibliography{main}
}
\end{document}
\fi

\ifx\experiments\undefined
\documentclass[10pt,twocolumn,letterpaper]{article}
\usepackage[review]{cvpr}      
\usepackage{graphicx}
\usepackage{amsmath}
\usepackage{amssymb}
\usepackage{booktabs}
\usepackage{makecell}
\usepackage{multirow}
\usepackage{CJKutf8}
\usepackage[pagebackref,breaklinks,colorlinks]{hyperref}
\usepackage[capitalize]{cleveref}
\crefname{section}{Sec.}{Secs.}
\Crefname{section}{Section}{Sections}
\Crefname{table}{Table}{Tables}
\crefname{table}{Tab.}{Tabs.}
\begin{document}
\fi


\begin{table}
\small
\begin{tabular*}{\hsize}{@{}@{\extracolsep{\fill}}p{3cm}p{1cm}p{1cm}p{0.50cm}p{0.7cm}@{}}
\toprule
Method&2DIoU(\%) &3DIoU(\%) & RMSE & $\delta_{1}$\\
\midrule

HorizonNet~\cite{horizon}         & 90.44 & 88.59 & 0.123 & 0.957  \\
LE$\mathrm{D}^{2}$-Net~\cite{led} & 90.36 & 88.49 & 0.124 & 0.955  \\
Ours [w/ Pure ViT]                   & 88.93 & 86.19 & 0.146 & 0.950  \\ 
Ours & \textbf{91.77} & \textbf{89.95} & \textbf{0.111} & \textbf{0.960} \\
\bottomrule
\end{tabular*}
\caption{Quantitative results of general layout estimation evaluated on ZInd\cite{cruz2021zillow} dataset.
}
\label{tab:zind}
\vspace{-4mm}
\end{table}

\ifx\experiments\undefined
{\small
\bibliographystyle{ieee_fullname}
\bibliography{main}
}
\end{document}
\fi

\paragraph{MatterportLayout} 
Evaluation of MatterportLayout\cite{zou2021manhattan} dataset is shown in \cref{tab:matterport}.
The results of LE$\mathrm{D}^{2}$-Net~\cite{led} are obtained from their official code\footnote{\label{led_url}\url{https://github.com/fuenwang/LED2-Net}} with re-training and re-evaluating by the standard evaluation metrics.
Moreover, we also report results with the post-processing of DuLa-Net\cite{yang2019dula} (denoted as ``Ours [w/ Post-proc]").
Our approach offers better performance than all other approaches with respect to 2DIoU, 3DIoU, and $\delta_{1}$.

We observe that similar approaches \cite{horizon, led, sun2021hohonet} of extracting the 1D feature sequence on equirectangular view are better than those using 2D convolutions\cite{zou2018layoutnet, yang2019dula}. 
In our opinion, Bi-LSTM\cite{schuster1997bidirectional, hochreiter1997long} and our SWG-Transformer are based on 1D horizontal feature sequence, which are better at establishing relations of the room layout. 

Qualitative comparisons are shown in \cref{fig:4_qualitative_comparison_mp3d}.
The first column shows that HorizonNet\cite{horizon} and LE$\mathrm{D}^{2}$-Net~\cite{led} predict discontinuous layouts at the  left and right borders of the panorama because they use Bi-LSTM\cite{schuster1997bidirectional, hochreiter1997long} to process the feature sequence and need to span the entire sequence while processing tokens at the first and last position.
However, our proposed SWG-Transformer treats tokens equally at all positions.
The second and third columns show that our approach better estimates the boundaries far from the camera center and those of complex room layouts.
Meanwhile, the visualizations of floor plans, normals, and gradients show that our approach offers better results by \emph{planar}-geometry awareness.
\vspace{-4mm}
\ifx\experiments\undefined
\documentclass[10pt,twocolumn,letterpaper]{article}
\usepackage[review]{cvpr}      
\usepackage{graphicx}
\usepackage{amsmath}
\usepackage{amssymb}
\usepackage{booktabs}
\usepackage{makecell}
\usepackage{multirow}
\usepackage{CJKutf8}
\usepackage[pagebackref,breaklinks,colorlinks]{hyperref}
\usepackage[capitalize]{cleveref}
\crefname{section}{Sec.}{Secs.}
\Crefname{section}{Section}{Sections}
\Crefname{table}{Table}{Tables}
\crefname{table}{Tab.}{Tabs.}
\begin{document}
\fi


\begin{figure*}
  \centering
  \includegraphics[width=1\linewidth]{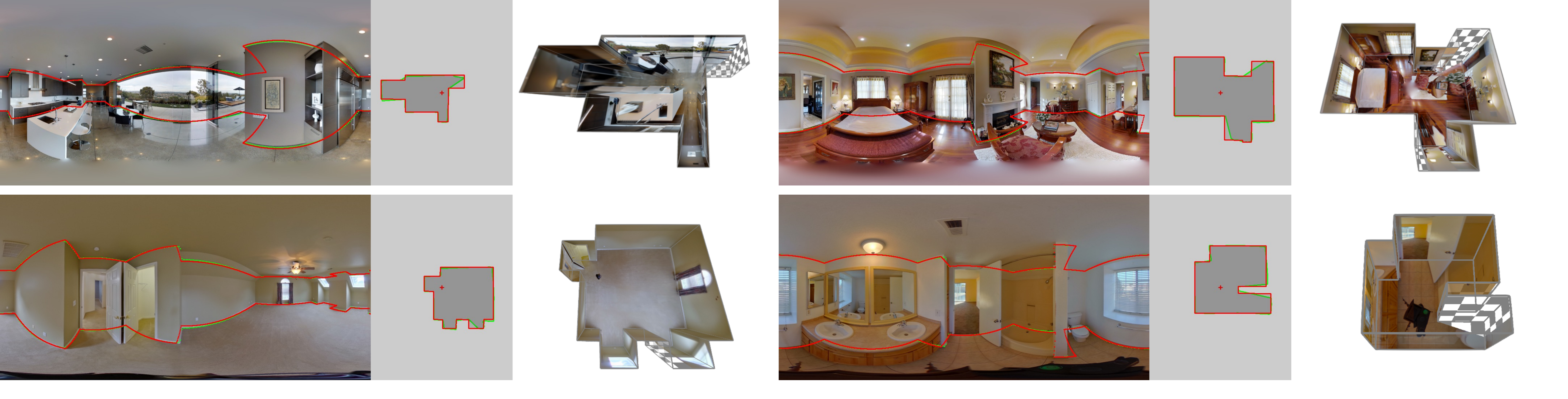}
  \caption{ The 3D visualization results of our approach on MatterportLayout\cite{zou2021manhattan} dataset (first row) and  ZInd\cite{cruz2021zillow} dataset (second row). The green lines are predicted boundaries by our network, and the red lines are results with post-processing of the prediction.
  } 
  \label{fig:4_visualizations}
   \vspace{-2mm}
\end{figure*}
\ifx\experiments\undefined
{\small
\bibliographystyle{ieee_fullname}
\bibliography{main}
}
\end{document}
\fi

\paragraph{ZInd}\label{zind_exp}
Evaluation on ZInd\cite{cruz2021zillow} dataset is shown in \cref{tab:zind}. 
The results of HorizonNet\cite{horizon} and LE$\mathrm{D}^{2}$-Net~\cite{led} are obtained from their official codes\footnoteref{led_url}\footnote{\url{https://github.com/sunset1995/HorizonNet}} with training and evaluating by the standard evaluation metrics. Our approach has higher accuracy than all other  approaches under all settings. 
Moreover, similar to the idea of ViT\cite{dosovitskiy2020image}, we split the panorama into patches by Patch Embedding\cite{dosovitskiy2020image} and feed them into our proposed SWG-Transformer (denoted as ``Ours [w/ Pure ViT]"). The results show that such ViT  architecture achieves comparable performance on the large dataset.

Qualitative comparisons are shown in \cref{fig:4_qualitative_comparison_zind}.
The first column shows that our SWG-Transformer can better process the left and right  borders of the panoramas.
The second column shows that our proposed \emph{omnidirectional}-geometry awareness has advantage on non-flat ceilings layouts since our approach is not affected by ceiling-boundary.
The third column shows that our approach performs better  with furniture occlusion than other approaches \cite{horizon, led}.

The 3D visualization results of our approach on MatterportLayout\cite{zou2021manhattan} dataset  and  ZInd\cite{cruz2021zillow} dataset are shown in \cref{fig:4_visualizations}. These examples show that our approach is effective in room layout estimation. See supplemental material for more qualitative results and quantitative results of different corners number and cross-dataset evaluation.

\subsection{Ablation Study}\label{section:ablation_study}
\ifx\experiments\undefined
\documentclass[10pt,twocolumn,letterpaper]{article}
\usepackage[review]{cvpr}      
\usepackage{graphicx}
\usepackage{amsmath}
\usepackage{amssymb}
\usepackage{booktabs}
\usepackage{makecell}
\usepackage{multirow}
\usepackage{CJKutf8}
\usepackage[pagebackref,breaklinks,colorlinks]{hyperref}
\usepackage[capitalize]{cleveref}
\crefname{section}{Sec.}{Secs.}
\Crefname{section}{Section}{Sections}
\Crefname{table}{Table}{Tables}
\crefname{table}{Tab.}{Tabs.}
\begin{document}
\fi


\begin{table}
\small
\begin{tabular*}{\hsize}{@{}@{\extracolsep{\fill}}p{3cm}p{1cm}p{1cm}p{0.50cm}p{0.7cm}@{}}
\toprule
Method&2DIoU(\%) &3DIoU(\%) & RMSE & $\delta_{1}$\\
\midrule
w/o Height & 82.82 & 80.44 & 0.205 & 0.945		 \\
w/o Nomal+Gradient& 84.24 & 81.86 & 0.196 & 0.954		 \\
w/o Gradient & 84.27 & 81.89 & \textbf{0.194 }& 0.954		 \\
\midrule
w/ \enspace Pure ViT  & 64.05 & 60.44 & 0.434 & 0.782		 \\
w/o Global Block & 83.02 & 80.40 & 0.212 & 0.947		 \\
w/ \enspace  Bi-LSTM & 83.98 & 81.32 & 0.201 & 0.950		 \\
w/o Window Block & 83.96 & 81.47 & 0.197 & \textbf{0.958}		 \\
\midrule
w/o PE & 83.78 & 81.50 & 0.197 & 0.951		 \\
w/ \enspace APE & 83.90 & 81.55 & 0.201 & 0.951		 \\
\midrule
Ours [Full] & \textbf{84.38} & \textbf{82.01} & \textbf{0.194} & 0.955 \\

\bottomrule
\end{tabular*}
\caption{Ablation study on MatterportLayout\cite{zou2021manhattan} dataset.}
\label{tab:ablation_study}
\vspace{-4mm}
\end{table}

\ifx\experiments\undefined
{\small
\bibliographystyle{ieee_fullname}
\bibliography{main}
}
\end{document}
\fi

Ablation study is shown in \cref{tab:ablation_study}. 
We reported results of the best performance of each configuration on the test split of MatterportLayout\cite{zou2021manhattan} dataset.
It should be noted that all experiments of ablation study select the best epoch in the test split. 
Thus, the results of ``Ours [full]" are higher than the corresponding quantitative results.

\vspace{-4mm}
\paragraph{Loss Function}
We replace the loss function in our approach with floor and ceiling horizon-depth errors like LE$\mathrm{D}^{2}$-Net~\cite{led} (denoted as ``w/o Height") and show that our proposed \emph{omnidirectional}-geometry aware loss of horizon-depth and room height significantly improves performance.
Moreover, our experiments without the normal and gradient errors (denoted as ``w/o Normal+Gradient" and ``w/o Gradient") show that our proposed \emph{planar}-geometry aware loss by normals and gradients of normals improves the performance.

\vspace{-4mm}
\paragraph{Network Architecture}
We experiment with ViT architecture (denoted as ``w/ Pure ViT") and show that ViT architecture does not achieve comparable performance in MatterportLayout\cite{zou2021manhattan} dataset. 
We argue that ViT architecture relies on large datasets like ZInd \cite{cruz2021zillow} to perform better.
Moreover, our experiments without Global Blocks or (Shifted) Window Blocks (denoted as ``w/o Global Block" and `w/o Window Block") demonstrate that using Window Blocks or Global Blocks alone leads to lower performance.
We replace SWG-Transformer with Bi-LSTM\cite{schuster1997bidirectional, hochreiter1997long} (denoted as ``w Bi-LSTM") and show that our SWG-Transformer offers better performance than Bi-LSTM.

\vspace{-4mm}
\paragraph{Position Embedding}
We experiment without position embedding (denoted as ``w/o PE") and only use absolute position embedding \cite{gehring2017convolutional} with learnable parameters (denoted as ``w/ APE").
These experiments show that absolute position embedding does not bring much improvement, but our designed relative position embedding offers the best performance.
We believe that since the contexts of panoramas constantly change in the horizontal direction, it is difficult to map the changes with a fixed absolute position embedding.

\ifx\allfiles\undefined
{\small
\bibliographystyle{ieee_fullname}
\bibliography{main}
}
\end{document}
\fi
\ifx\allfiles\undefined
\documentclass[10pt,twocolumn,letterpaper]{article}
\usepackage[review]{cvpr}      
\usepackage{graphicx}
\usepackage{amsmath}
\usepackage{amssymb}
\usepackage{booktabs}
\usepackage{makecell}
\usepackage{multirow}
\usepackage{CJKutf8}
\usepackage[pagebackref,breaklinks,colorlinks]{hyperref}
\usepackage[capitalize]{cleveref}
\crefname{section}{Sec.}{Secs.}
\Crefname{section}{Section}{Sections}
\Crefname{table}{Table}{Tables}
\crefname{table}{Tab.}{Tabs.}
\makeatletter
\newcommand\footnoteref[1]{\protected@xdef\@thefnmark{\ref{#1}}\@footnotemark}
\makeatother

\begin{document}
\fi

\def\experiments{}
\section{Conclusions}\label{sec:conclusions}
In this paper, we proposed an efficient model, LGT-Net, for 3D room layout estimation. Horizon-depth and room height  offer omnidirectional geometry awareness. Normals and gradients of normals offer planar geometry awareness.
Moreover, the proposed SWG-Transformer with the noval relative position embedding  can better establish the local and global geometry relations of the room layout. We evaluate our approach on both cuboid and general datasets and show better performance than the baselines.
\vspace{-4mm}
\paragraph{Acknowledgement} This work is supported by the National Natural Science Foundation of China 
under Project 61175116.
\ifx\allfiles\undefined
{\small
\bibliographystyle{ieee_fullname}
\bibliography{main}
}
\end{document}
\fi

{\small
\bibliographystyle{ieee_fullname}
\bibliography{main}
}

\end{document}